\documentclass[10pt,twocolumn,letterpaper]{article}
\usepackage[accsupp]{axessibility}
\usepackage{cvpr}              

\usepackage{amsmath}
\usepackage{amssymb}
\usepackage{kotex}
\usepackage{tikz}

\usepackage{mwe}
\usepackage{algorithm}
\usepackage{algpseudocode}

\usepackage{multirow}
\usepackage{xcolor}
\usepackage{booktabs}
\usepackage{colortbl}
\usepackage{makecell}
\usepackage{subcaption}
\newif \ifauthorcomment
\authorcommenttrue

\newcounter{comments}

\definecolor{lightgray}{HTML}{E7E6E6}
\newcommand{\PP}[1]{
\noindent{\textbf{#1}}
}

\algrenewcommand\algorithmiccomment[1]{\textbf{$\triangleright$ #1}\hfill}

\newcommand{\ours}{\textit{\textbf{Phantom}}}

\newcommand{\hide}[1]{}
\usepackage{setspace}
\usepackage{microtype}

\definecolor{cvprblue}{rgb}{0.21,0.49,0.74}
\usepackage[pagebackref,breaklinks,colorlinks,allcolors=cvprblue]{hyperref}

\def\paperID{45806} 
\def\confName{CVPR}
\def\confYear{2026}

\title{\ours: A Unified Face-Swap Deepfake Protection Framework\\with Latent and Spatial Constraints}

\author{Jungkon Kim\thanks{Corresponding author} \quad Cheolseung Jung \quad Jong-Min Choi \quad Juseong Lee\\
Samsung Electronics, AI Platform Center\\
{\tt\small \{jungkon.kim, cs.jung, jminl.choi, jooseong.lee\}@samsung.com}
}

\begin{document}
\maketitle
\begin{abstract}

Face-swapping deepfakes pose an escalating threat to personal privacy by enabling unauthorized identity manipulation. While adversarial approaches have demonstrated success against black-box face recognition (FR) models, their applicability to face-swapping scenarios remains underexplored. In particular, reliance on fixed or random targets yields ambiguous latent guidance, and the lack of explicit spatial constraints causes perturbations to spill into identity-irrelevant regions. These issues are further exacerbated by identity–style disentanglement, which suppresses adversarial signals during deepfake generation. In this paper, we present $\ours$, a unified face-swap deepfake protection framework that jointly constrains perturbations in latent and spatial domains. $\ours$ adaptively synthesizes identity-shifted yet attribute-preserving targets to guide identity-aware latent optimization, and applies masked perturbations confined to semantically relevant facial regions. Extensive experiments on state-of-the-art face-swapping deepfakes demonstrate that $\ours$ improves protection success rates in dodging scenarios by 27.8\%, 25.6\%, and 16.6\% on UniFace, INSwapper, and SimSwap, respectively, while also enhancing visual quality. Furthermore, $\ours$ generalizes to impersonation scenario, yielding up to 10.2\% higher protection while improving perceptual fidelity. These results underscore the effectiveness of jointly leveraging latent and spatial constraints for robust and coherent facial privacy protection.
\end{abstract}

\section{Introduction}
\label{sec:intro}

\begin{figure}[t!]
    \centering
    \includegraphics[width=\linewidth]{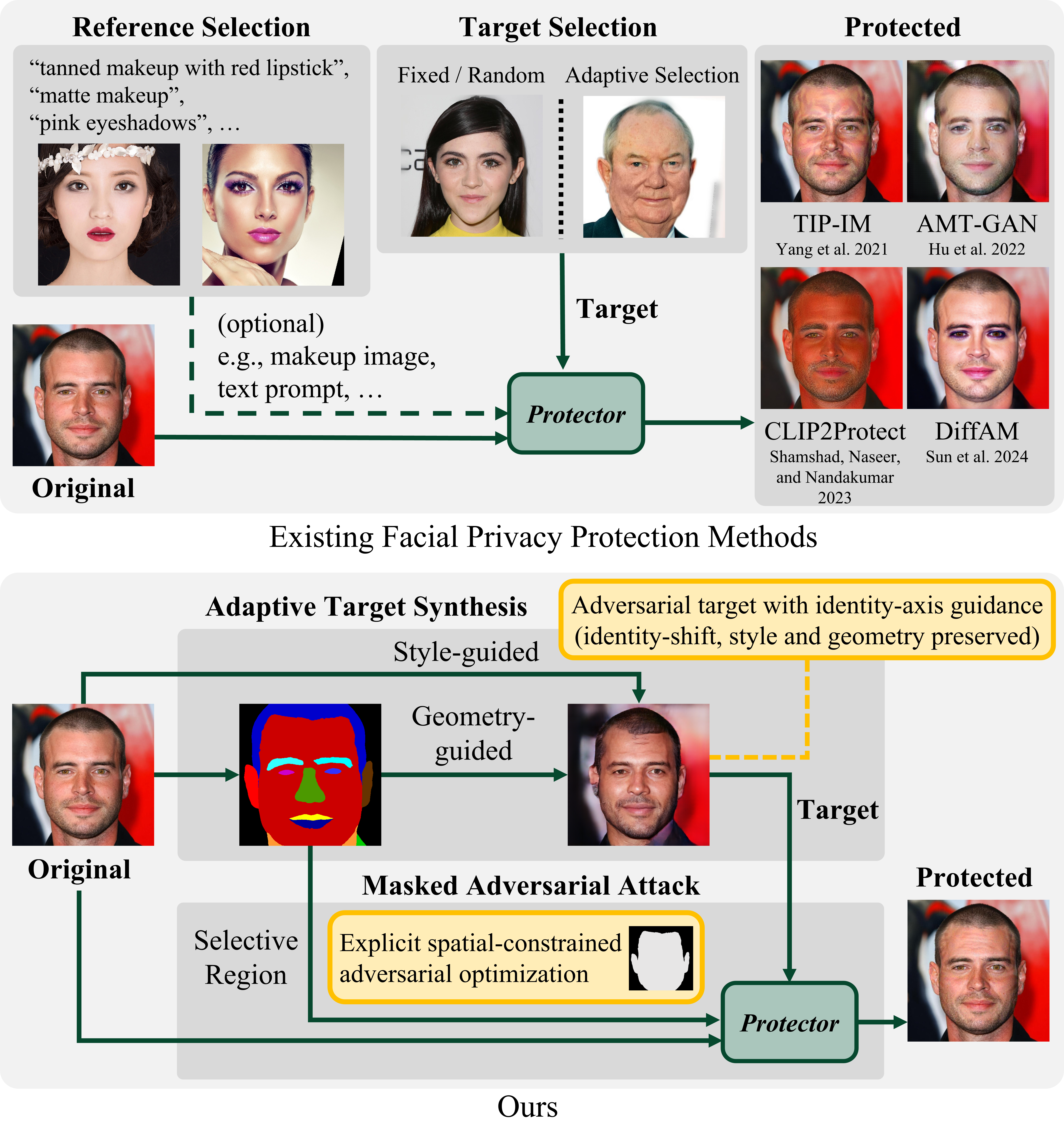}
    \caption{\textls[-10]{Core idea comparison. Existing methods define perturbations using selected targets and apply noise broadly, often ignoring spatial relevance and identity alignment. Our method synthesizes attribute-preserving targets that guide updates along identity-aware axes in latent space, and explicitly constrains perturbations to identity-related regions, enabling compact and visually coherent protection.}}
    \label{fig:intro}
\end{figure}

The advancement of deepfake technology raises significant privacy concerns, particularly as face-swapping deepfakes facilitate unauthorized identity manipulation. To counteract this threat, many researches~\cite{Cozzolino_2021_ICCV,Dong_2022_CVPR,dong2023implicit} have proposed for detecting deepfake-generated content. However, these retrospective approaches can only mitigate damage after attacks have occurred. To overcome this, recent research has increasingly focused on proactive strategies to disrupt deepfake generation.

Adversarial perturbation has emerged as a proactive protection strategy, disrupting face recognition via either patch-based~\cite{hu2021naturalistic,xiao2021improving,guesmi2024dap} or noise-based~\cite{goodfellow2014explaining,madry2017towards,dong2018boosting,dong2019evading, yang2021towards} methods. Patch-based approaches are lightweight and modular but tend to be visually conspicuous and spatially fixed, rendering them vulnerable to trivial removal~\cite{jing2024pad}. Noise-based methods typically achieve strong protection by applying perturbations over the entire face bounding box. However, this global application leads to inefficient updates and perceptual degradation, as perturbations are dispersed across identity-irrelevant regions. A recent work~\cite{chow2024personalized} proposes a per-user strategy that trains identity-specific masks from multiple images. While effective, it requires offline optimization and user-level data, limiting its applicability in per-image protection we target. Makeup-based methods~\cite{hu2022protecting,shamshad2023clip2protect,sun2024diffam,liu2025recoverable} embed perturbations into visually plausible styles using reference-guided generation. While producing visually natural transformations, these methods often require exaggerated modifications and offer limited spatial control to ensure sufficient protection. Moreover, as most are optimized for impersonation, their applicability to dodging-oriented face-swap deepfake protection remains relatively underexplored.

Another critical limitation of existing methods lies in the definition of perturbation directions in the embedding space. Most prior works define identity-shifting directions using fixed or randomly selected target images~\cite{yang2021towards,hu2022protecting,shamshad2023clip2protect,sun2024diffam}, which entangle identity with confounding attributes such as pose or texture. This often results in arbitrarily maximizing latent distances, leading to ambiguous or suboptimal updates that compromise visual quality or fail to ensure sufficient protection.
Recent work by \citet{liu2025recoverable} proposes an adaptive target selection strategy based on landmark similarity, which partially constrains geometric variation. However, it still struggles to isolate identity-specific changes, often guiding perturbations along directions that include residual non-identity attributes, thereby limiting its effectiveness.

In this paper, we propose $\ours$, a unified framework for face-swap deepfake protection that jointly performs adversarial optimization under latent and spatial constraints (See Fig.~\ref{fig:intro}.). 


First, we introduce adaptive target synthesis strategy, which generates identity-shifted yet attribute-preserving target faces to define a semantic direction in latent space for adversarial guidance. Unlike prior works~\cite{yang2021towards,hu2022protecting,shamshad2023clip2protect,sun2024diffam,liu2025recoverable} that treat target identities as final objectives, we reinterpret the synthesized image as a means of defining a semantically meaningful identity-discriminative direction in the latent space intertwining geometry, style, and identity cues. Specifically, we extract a semantic mask from the input face and reapply the original style onto it to generate a modified image that retains structural and stylistic consistency while inducing subtle identity shifts. With this adaptively synthesized target, we guide adversarial perturbation generation along identity-sensitive directions, enabling effective identity suppression without unintended attribute distortion.

Second, we formulate a masked adversarial attack based on spatial-constrained optimization. Specifically, we constrain gradient updates to semantically important facial regions using segmentation masks. By explicitly guiding perturbations toward identity-aware regions, this strategy ensures that noise is both compact and effective, minimizing visual artifacts while enhancing protection robustness.

Extensive experiments on CelebA-HQ~\cite{karras2017progressive} and LADN~\cite{gu2019ladn} datasets validate the effectiveness of $\ours$ across both dodging and impersonation scenarios. In dodging scenarios, which are our primary focus for defending against unauthorized deepfakes, $\ours$ achieves state-of-the-art protection against black-box FR models while preserving visual quality. Furthermore, we evaluate whether such protection is valid against face-swapping deepfakes, where adversarial signals including those in stylized or unconstrained formats are often discarded due to the disentanglement of identity and style representations. $\ours$ achieves protection success rate improvements of 27.8\%, 25.6\%, and 16.6\% over state-of-the-art methods on UniFace~\cite{xu2022designing}, INSwapper~\cite{InsightFace}, and SimSwap~\cite{chen2020simswap}, respectively, indicating strong robustness not only against FR models but also in face-swapping deepfake environments.

Moreover, we demonstrate that $\ours$ generalizes to impersonation scenarios that induce recognition as a target identity. While the adaptive synthesis module is deactivated in this setting, our spatial optimization remains effective and yields strong protection along with enhanced visual fidelity. 

In summary, our contributions are as follows:
\begin{itemize}
    \item We present $\ours$, a unified face-swap deepfake protection framework that combines latent- and spatial-constrained optimization, jointly addressing both directional and regional limitations of prior methods.
    \item We propose a latent-constrained adversarial optimization strategy via adaptive target synthesis that produces identity-shifted yet structure-preserving targets, offering coherent directional guidance in embedding space.
    \item We introduce a spatial-constrained adversarial optimization strategy using semantically masked gradient updates, enabling compact and visually aligned perturbations for enhanced robustness.
    \item $\ours$ outperforms existing methods in protecting against black-box FR models and generalizes effectively to face-swapping deepfakes while improving visual quality. Notably, it achieves up to 98.2\% protection success against face-swapping attacks, surpassing prior approaches by at least 16.6\%, and also shows competitive performance in impersonation scenarios.
\end{itemize}

\section{Related Work}
\label{sec:rw}
\begin{figure*}[t]
    \centering
    \includegraphics[width=0.9\linewidth]{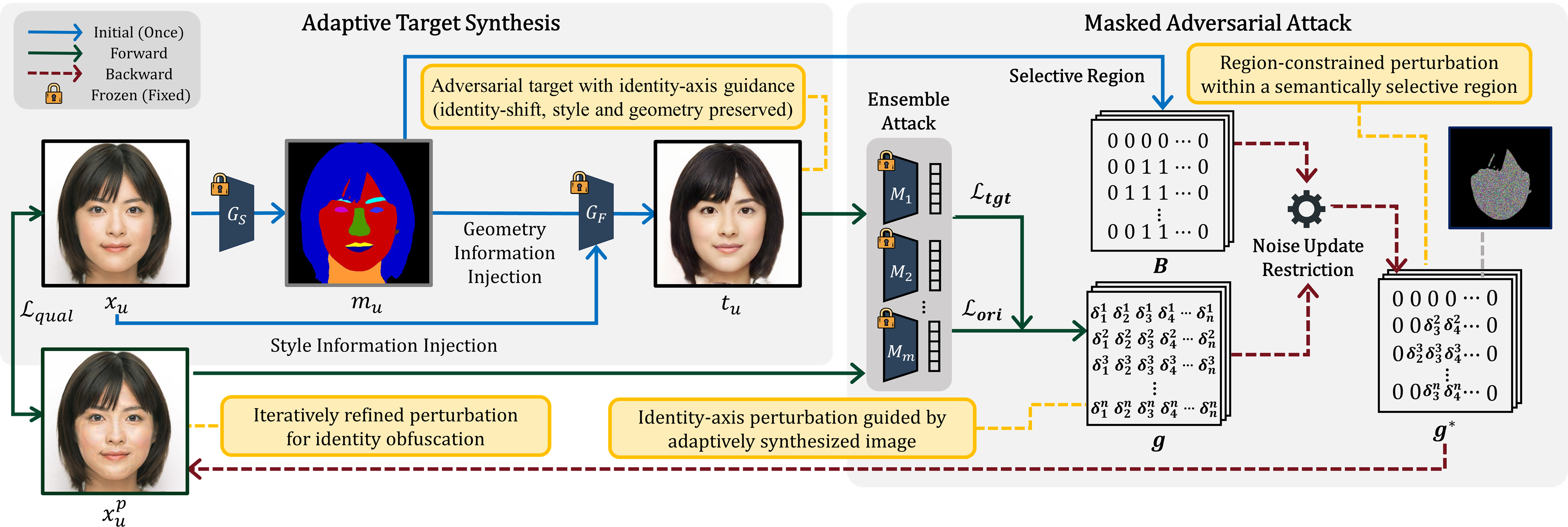}
    \caption{Overview of $\ours$, our unified face-swap deepfake protection framework.}
    \label{fig:overview}
\end{figure*}
\subsection{Adversarial Attacks on Face Recognition}

Adversarial attacks are categorized as targeted (impersonation) or non-targeted (dodging). Dodging attacks attempt to conceal the true identity of the source image, forcing recognition models to misclassify it as an unknown or incorrect person. 
In contrast, impersonation attacks aim to manipulate the image such that it is recognized as a specific target identity, thereby achieving targeted identity transfer.

Noise-based methods typically apply imperceptible perturbations over the entire facial region~\cite{goodfellow2014explaining, madry2017towards}, but often suffer from limited visual quality. Subsequent works have improved robustness through momentum~\cite{dong2018boosting} and translation-invariant strategies~\cite{dong2019evading}. TIP-IM~\cite{yang2021towards} employs targeted optimization to generate adversarial identity masks effective against black-box FR models. However, this approach introduces perceptible noise within the face bounding box, thereby degrading perceptual quality.

Makeup-based approaches~\cite{hu2022protecting, shamshad2023clip2protect, sun2024diffam, liu2025recoverable} embed adversarial perturbations into visually plausible styles using reference images or text prompts. Although \citet{shamshad2023clip2protect} addresses both impersonation and dodging, most methods are optimized for impersonation and remain largely unexplored under face-swapping deepfake scenarios. Moreover, these methods offer limited spatial control and high computational cost.

\subsection{Face Manipulation}

Face manipulation models aim to generate realistic facial images while controlling attributes such as expression or style. Recent approaches create and leverage semantic segmentation masks to enable structure-aware generation~\cite{lee2020maskgan, durall2021facialgan}. They synthesize faces conditioned on semantic masks and reference images, allowing localized style transfer. While these models attempt to preserve identity, slight changes can arise during style adaptation. \citet{huang2023collaborative} extends this direction by generating identity-diverse images from segmentation masks in a structure-guided manner and optional text prompts for style.


\subsection{Face-Swapping Deepfakes}

Face-swapping deepfakes synthesize realistic identity manipulations by disentangling identity features from style (e.g., pose, texture) and recombining them across subjects. Earlier methods such as FaceShifter~\cite{li1912faceshifter} and SimSwap~\cite{chen2020simswap} employ encoder–decoder architectures with FR-guided identity injection, while recent models like UniFace~\cite{xu2022designing} introduce explicit identity and style branches to enhance controllability and fidelity.

\section{Methodology}

We propose $\ours$, a structured adversarial framework for facial privacy protection that proactively defends against black-box FR and face-swapping deepfakes. As illustrated in Fig.~\ref{fig:overview}, $\ours$ integrates two complementary optimization strategies: a latent-constrained adaptive target synthesis strategy that guides embedding shifts along identity-discriminative directions, and a spatial-constrained masked adversarial attack that restricts perturbations to semantically meaningful regions. Together, these components enable spatially focused and identity-aware perturbations while maintaining high visual fidelity.

\subsection{Problem Formulation}


We consider a practical threat scenario where user face images are collected from public sources, such as social media or online platforms, and are exploited without consent. Such images may be used by adversaries for unauthorized face recognition (FR) or identity manipulation via face-swapping deepfakes. Our goal is to proactively defend against both threats by generating visually plausible but identity-disruptive variants.

Let $x_u$ and $t_u$ denote the original and target face images of user $u$, respectively, and let $x_u^p$ be its protected version. Given a black-box FR model $f(\cdot)$, we denote the embedding (i.e., the latent representation) of an image $x$ as $\mathbf{z}_x = f(x)$. Our goal is to generate $x_u^p$ such that identity inference is suppressed by disrupting $\mathbf{z}_{x_u^p}$, while remaining visually similar to $x_u$. To this end, we define a unified objective that supports both dodging (maximizing deviation from $\mathbf{z}_{x_u}$) and impersonation (minimizing distance to $\mathbf{z}_{t_u}$), formulated as:

\begin{equation}\label{eq:general_problem}
\begin{array}{c}
\displaystyle \max_{x_u^p} \; \mathcal{L}_{adv} = D(\mathbf{z}_{x_u^p}, \mathbf{z}_{x_u}) - D(\mathbf{z}_{x_u^p}, \mathbf{z}_{t_u}) \\[0.8em]
\displaystyle \text{s.t.} \quad L_p(x_u^p, x_u) \leq \epsilon
\end{array}
\end{equation}

\noindent
Here, $D(\cdot)$ is a similarity metric (e.g., cosine or Euclidean), and $L_p(\cdot)$ denotes the perceptual distance between $x_u^p$ and $x_u$, bounded by a threshold $\epsilon$ to ensure imperceptibility.

\subsection{Adaptive Target Synthesis}

\begin{figure}[!hp]
    \centering
    \begin{minipage}{0.49\columnwidth}
        \centering
        \includegraphics[width=\columnwidth]{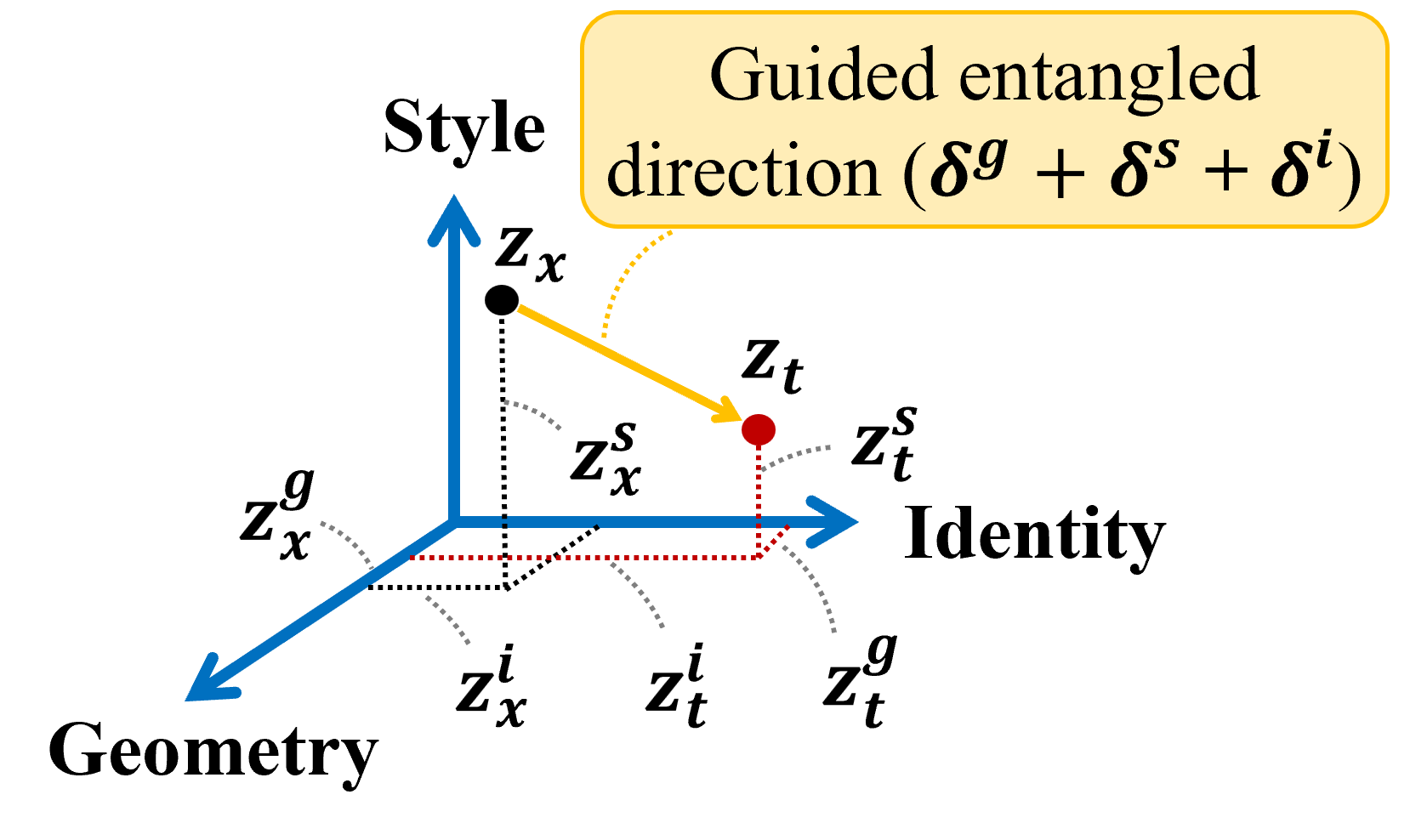}
        \subcaption{selection-based.}
        \label{fig:left_origin}
    \end{minipage}\hfill
    \begin{minipage}{0.49\columnwidth}
        \centering
        \includegraphics[width=\columnwidth]{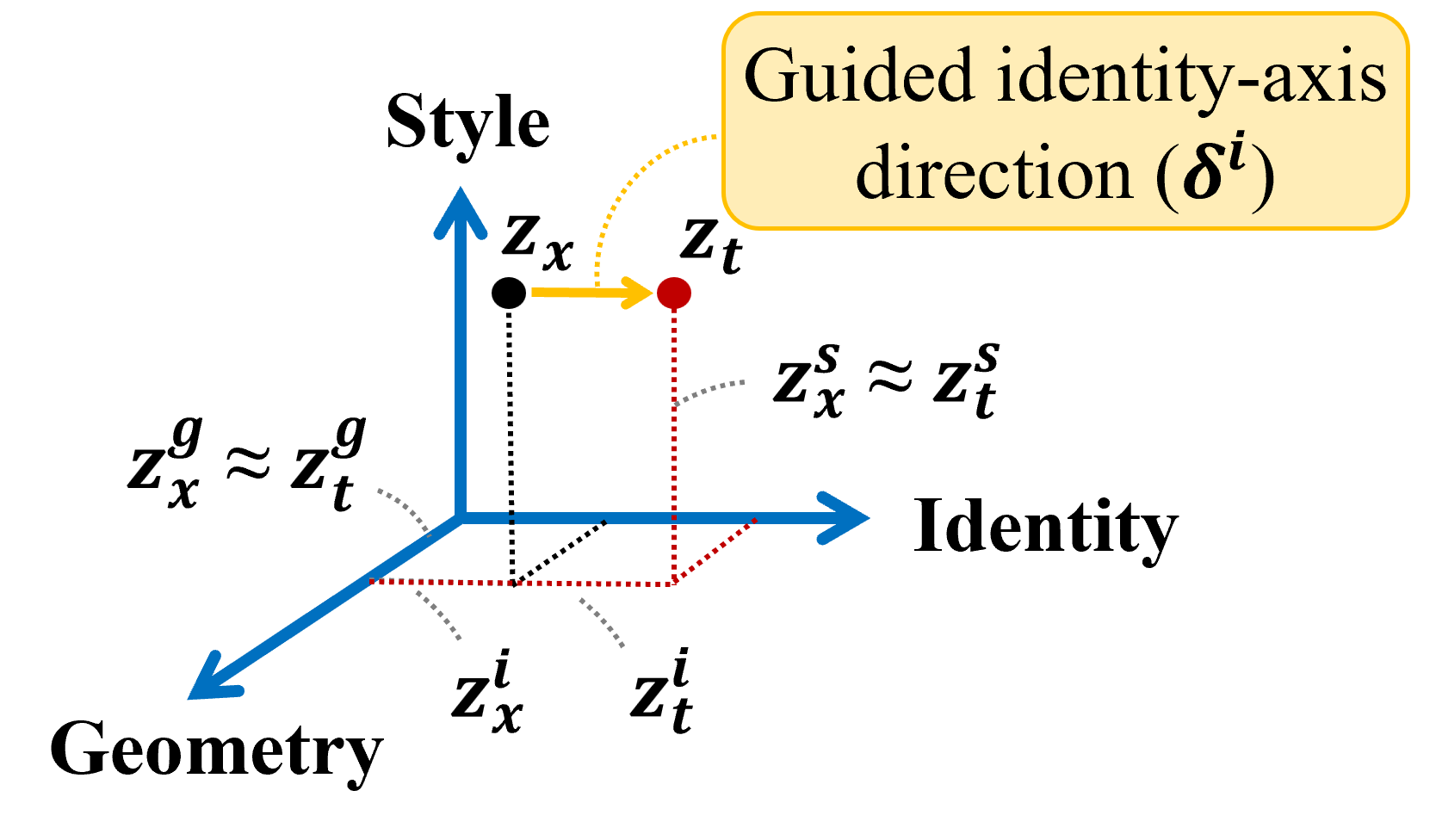}
        \subcaption{(proposed) synthesis-based.}
        \label{fig:right_proposed}
    \end{minipage}
    
    \caption{Comparison of guided perturbation directions in latent space across target strategies.}
    \label{fig:identity_axis}
\end{figure}

Motivated by the entangled nature of identity, style, and geometry in the latent space of FR models, we seek to isolate identity-discriminative directions while preserving non-identity factors. To this end, we synthesize a target image $t$ that aligns with the source $x$ in style and geometry but differs subtly in identity. As depicted in Fig.~\ref{fig:identity_axis}, conventional target selection strategies (e.g., random or fixed) yield perturbations that traverse mixed directions in the latent space, including identity ($\delta^i = z_t^i - z_x^i$), style ($\delta^s = z_t^s - z_x^s$), and geometry ($\delta^g = z_t^g - z_x^g$), which are not optimized solely for identity shift. In contrast, our synthesized $t$ constrains $\delta^s$ and $\delta^g$ near zero by construction, thereby encouraging perturbations aligned predominantly along the identity axis $\delta^i$.
Note that visual similarity between $x$ and $t$ is not a concern; rather, it is expected and acceptable, as the objective is not to maximize visual divergence but to extract a semantic direction $\delta^i$ for identity-axis perturbation. Unlike unconstrained directions that may entangle identity with other attributes, this targeted guidance allows $\ours$ to suppress identity-discriminative features more effectively with minimal perturbations.

To synthesize a target $t$ disentangled from $x$ only in terms of identity, we repurpose face manipulation models. Given an input $x$, we first extract a semantic mask $m = G_S(x)$ using a segmentation model $G_S$. This mask is reused without modification to retain spatial layout. A mask-guided face generation model $G_F$ then combines $x$ and $m$ to produce the synthesized target:
\begin{equation}\label{eq:atsyn}
t \gets G_F(x, G_S(x))
\end{equation}
Although $G_F$ is originally intended for tasks such as retouching or style transfer, it introduces subtle identity shifts while maintaining structure. We reinterpret these changes as semantically aligned traversals in latent space that serve as directional anchors for identity-axis optimization.

\subsection{Masked Adversarial Attack}

\PP{Definition of Selective Region.}
We identify an optimal perturbation region that maximizes protection against both FR models and face-swapping while minimizing perceptual degradation.
Since the spatial placement of adversarial perturbation critically impacts both effectiveness and visual quality, selecting the effective region is essential.

To this end, we empirically evaluate various perturbation scopes including the entire image, full face region, non-face areas (e.g., background, hair), and localized facial components (e.g., eyes, nose, mouth). As detailed in the Supplementary Material, perturbations confined to the face region yield the best trade-off, offering strong protection with minimal artifacts. In contrast, applying perturbations to irrelevant or overly localized regions reduces effectiveness or introduces visible distortions. Based on this analysis, we adopt the face region as our selective region for adversarial optimization.

\PP{Explicit Region-Constrained Optimization.}
We formulate a proactive adversarial attack strategy that suppresses identity features in facial images, thereby degrading their utility for downstream recognition or swapping tasks. $\ours$ combines adaptive target synthesis with spatially constrained perturbation to generate identity-obfuscating perturbations confined to semantically meaningful regions. This enables effective protection under black-box conditions, including unseen FR models or deepfake generators, while preserving image fidelity. The overall objective consists of identity-oriented and quality-preserving terms:
\begin{equation}\label{eq:loss}
\begin{aligned}
\mathcal{L}_{total} = \lambda_{id}\lambda_{ori}\mathcal{L}_{ori} + \lambda_{id}\lambda_{tgt}\mathcal{L}_{tgt} + \lambda_{qual}\mathcal{L}_{qual}
\end{aligned}
\end{equation}
$\mathcal{L}_{ori}$ and $\mathcal{L}_{tgt}$ represent cosine similarity losses against the original and target embeddings, computed using an ensemble of surrogate FR models to improve transferability. $\mathcal{L}_{qual}$ enforces perceptual fidelity using pixel-wise MSE loss, and helps bound the perturbation strength by limiting the maximum per-pixel deviation to $\epsilon$.

We apply per-iteration clamping using a binary mask to constrain adversarial updates within the predefined region. Specifically, a binary mask $B_u \in {0,1}^{H \times W}$ is constructed based on the selected facial region. During optimization, we compute the gradient $g$ of the total loss with respect to the adversarial perturbations $n_{adv}$ and apply a Hadamard product with the mask:
\begin{equation}\label{eq:hadmard}
\begin{aligned}
g^* = B_u \odot g, \quad \text{s.t.} \quad g = \nabla_{n_{adv}} \mathcal{L}_{total}
\end{aligned}
\end{equation}
This yields compact, task-aware perturbations that are both visually natural and effective against unseen FR models or deepfake generators.

\PP{Generalization to Impersonation Scenario.}
While $\ours$ primarily targets dodging scenario (e.g., avoiding unauthorized FR models or deepfakes), it can be naturally extended to impersonation settings. In such cases, adaptive target synthesis is not required. Instead, a specific identity image can be directly selected as $t_u$ to steer the embedding toward a desired target identity. The rest of the masked adversarial optimization framework remains unchanged.

The full algorithm of $\ours$ is provided in the Supplementary Material.

\section{Experiments}
\label{sec:exp}

\begin{figure*}[!htbp]
    \centering
    \includegraphics[width=\linewidth]{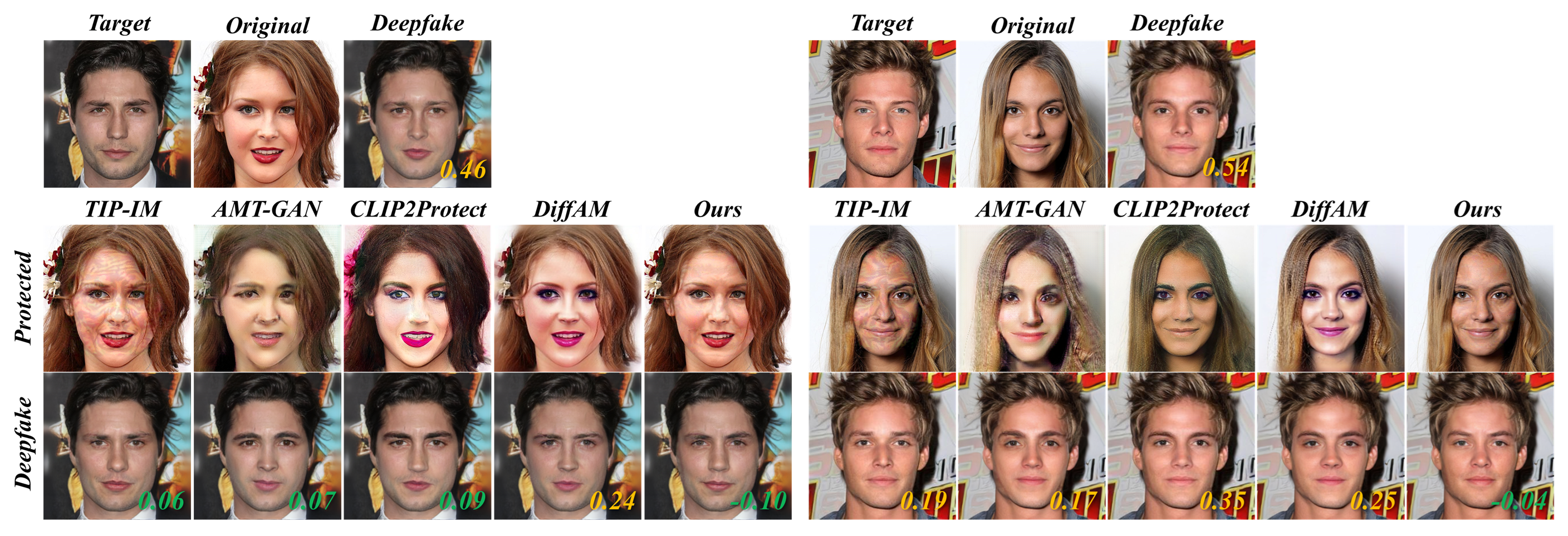}
    \caption{Deepfake results with UniFace. Numbers represent the cosine similarity on IR152 between deepfake images and the original images. Yellow and Green indicate protection failure and success based on a threshold of 0.167 at FAR@0.01.}
    \label{fig:df_result}
\end{figure*}

\subsection{Experimental Settings}~\label{subsec:exp_setting}

\PP{Implementation Details.} We set $iter{=}200$, $\epsilon{=}16$, and use the Adam optimizer with a learning rate of $5\mathrm{e}{-4}$, $\beta_1{=}0.9$, and $\beta_2{=}0.999$. Loss weights are configured as $(\lambda_{id}, \lambda_{qual}){=}(1.0, 1.0)$; for dodging, $(\lambda_{ori}, \lambda_{tgt}){=}(1.0, 0.5)$, and for impersonation, $(0.0, 1.0)$. We use $G_S$ and $G_F$ from~\citet{lee2020maskgan}, which preserve both style and geometry. To assess robustness under different synthesis characteristics, we also evaluate $G_F$ from~\citet{huang2023collaborative}, which preserves geometry only. 

\PP{Datasets.} 
Following~\cite{hu2022protecting, sun2024diffam,liu2025recoverable}, we use the CelebA-HQ~\cite{karras2017progressive} and LADN~\cite{gu2019ladn} datasets. For CelebA-HQ, we select 1,000 images from~\cite{hu2022protecting} and use 930 image pairs with corresponding counterparts. One pair is randomly selected per image based on official pair information. For LADN, we evaluate on the same 334 images used by existing methods to ensure a fair comparison.

\PP{Target FR Models.} 
Following~\cite{hu2022protecting,shamshad2023clip2protect,sun2024diffam,liu2025recoverable}, we evaluate $\ours$ against four FR models with different backbone architectures: IRSE50~\cite{hu2018squeeze}, IR152~\cite{deng2019arcface}, FaceNet~\cite{schroff2015facenet}, and MobileFace~\cite{chen2018mobilefacenets}.
Before feeding images into these FR models, we detect faces using RetinaFace~\cite{deng2020retinaface} implemented in the InsightFace~\cite{InsightFace}. 

\PP{Deepfakes.} 
We evaluate our approach against face-swapping deepfake attacks using three state-of-the-art tools: INSwapper~\cite{InsightFace}, deployed in Picsi.Ai~\cite{Picsi.Ai}, UniFace~\cite{xu2022designing}, and SimSwap~\cite{chen2020simswap}. Each tool utilizes a different pretrained swapper model with unique resizing configurations in pre-processing.
See the Supplementary Material for details.

\PP{Compared Methods.}
We compare our approach against representative benchmark approaches. Perturbation-based methods include MI-FGSM~\cite{dong2018boosting}, TI-DIM~\cite{dong2019evading}, and TIP-IM~\cite{yang2021towards}. Makeup-based methods include AMT-GAN~\cite{hu2022protecting}, CLIP2Protect~\cite{shamshad2023clip2protect}, DiffAM~\cite{sun2024diffam}, and RMT-GAN~\cite{liu2025recoverable}.
All methods adopt the ensemble strategy from~\cite{hu2022protecting} for fair comparison.
For the dodging scenario, methods without explicit support for dodging-specific optimization (e.g., AMT-GAN~\cite{hu2022protecting}, DiffAM~\cite{sun2024diffam}) are evaluated using their official implementations and recommended hyperparameters from the original papers. Details in the Supplementary Material.

\PP{Evaluation Metrics.}
To assess protection effectiveness, we report Rank-1/5 protection (R1-P/R5-P) and Protection Success Rate (PSR) under FAR@0.01. R1-P/R5-P measure how well protection methods prevent the ground-truth pair from appearing in the top-N predictions on the FR model, while PSR quantifies how often a deepfake image fails to match the true identity under a fixed FAR. For PSR, the model-specific thresholds are 0.241 (IRSE50), 0.167 (IR152), 0.409 (FaceNet), and 0.302 (MobileFace). The Attack Success Rate (ASR) is defined as $(1 - \text{PSR})$ for a target identity. For image quality, we use PSNR, SSIM~\cite{wang2004image}, and FID~\cite{heusel2017gans}. Higher R1/R5 and PSR indicate stronger protection; higher PSNR and SSIM, and lower FID indicate better visual quality.

In all experiments, we evaluated on an NVIDIA A100 GPU with 80GB of memory.

\subsection{Dodging Scenario Evaluation}~\label{subsec:dodging}

\PP{Evaluation against unseen FR models.}
We evaluate $\ours$ in a black-box setting using four unseen FR models. As shown in Tab.~\ref{tbl:ranked_dodging}, $\ours$ achieves an average protection of 95.0\% in R1-P and 89.6\% in R5-P, outperforming existing methods by 9.1\% and 18.2\%, respectively. This confirms strong identity obfuscation across diverse FR architectures. AMT-GAN and DiffAM show limited performance due to their impersonation-oriented design. Tab.~\ref{tbl:qual_dodging} reports SSIM, PSNR, and FID scores, showing that $\ours$ preserves perceptual quality (SSIM: 0.973, PSNR: 33.646, FID: 18.705) while maintaining strong protection.
\begin{table}[!]
\centering
\caption{R1-P/R5-P performance (\%) comparison of different methods on CelebA-HQ dataset. For each columns, the other three FR models are used as surrogates to generate the protected images.}
\label{tbl:ranked_dodging}
\resizebox{\linewidth}{!}{
\begin{tabular}{l||cc||cc||cc||cc}
\toprule
& \multicolumn{2}{c||}{\textbf{IRSE50}} & \multicolumn{2}{c||}{\textbf{IR152}} & \multicolumn{2}{c||}{\textbf{FaceNet}} & \multicolumn{2}{c}{\textbf{MobileFace}}  \\ 
\multirow{-2}{*}{\textbf{Method}} & R1-P & R5-P & R1-P & R5-P & R1-P & R5-P & R1-P & R5-P \\ 
\bottomrule

Clean   &  07.3&  04.4&  04.0&  03.3&  16.2&  07.7&  18.0&  10.8\\
TIP-IM~\cite{yang2021towards} & 84.5 & \underline{73.3} & 65.2 & 60.0 & 82.4 & 66.6 & 81.9 & 68.3 \\
AMT-GAN~\cite{hu2022protecting}  &  71.1&  52.8& 65.2&  47.3&  \underline{87.3} &  \underline{70.9}&  76.5&  59.4\\
CLIP2Protect (CVPR 2023) & \underline{87.7} & \underline{73.3} & \underline{78.4} & \underline{61.3} & 83.3 & 65.8 & \underline{94.5} & \underline{85.2}\\
DiffAM~\cite{sun2024diffam}  & 68.9&  42.1&  44.5&  25.6&  78.0&  55.1&  80.5&  57.4 \\
Ours $(\lambda_{ori}{ = }1.0, \lambda_{tgt}{ = }0.5)$ & \textbf{95.8} & \textbf{91.9} & \textbf{91.6} & \textbf{85.4} & \textbf{93.9} & \textbf{83.9} & \textbf{98.8} & \textbf{97.3} \\
\bottomrule
\end{tabular}
}
\end{table}


\begin{table}[!]
\centering
\footnotesize
\caption{Perceptual Quality under Dodging Setting}
\label{tbl:qual_dodging}

\resizebox{\linewidth}{!}{%
\begin{tabular}{l||ccc}
\toprule
Method & SSIM ($\uparrow$) & PSNR ($\uparrow$) & FID ($\downarrow$)  \\
\midrule

TIP-IM~\cite{yang2021towards} &  0.884& 30.715 & 30.858\\
AMT-GAN~\cite{hu2022protecting} &  0.787&  19.505& 34.441\\
CLIP2Protect~\cite{shamshad2023clip2protect} &  0.570&  16.904& 39.890\\
DiffAM~\cite{sun2024diffam} &  0.874& 20.526 &26.102\\
\midrule
 Ours $(\lambda_{ori}{ = }1.0, \lambda_{tgt}{ = }0.5)$ &  \textbf{0.973}&  \textbf{33.646}& \textbf{18.705}\\
\bottomrule
\end{tabular}%
}
\end{table}
\begin{table*}[!htbp]
\centering
\caption{ Evaluation of PSR for dodging (non-targeted) scenario at FAR@0.01 against state-of-the-art face-swapping deepfakes: UniFace~\cite{xu2022designing}, INSwapper~\cite{InsightFace}, and SimSwap~\cite{chen2020simswap}.
For each evaluated FR model, the other three FR models are used as surrogates to generate the protected images.}
\label{tbl:deepfake_result}
\resizebox{\textwidth}{!}{%
\begin{tabular}{l|c||cccc|c||cccc|c}
\toprule

\multirow{2}{*}{\textbf{Method}} & & \multicolumn{5}{c||}{\textbf{CelebA-HQ}} & \multicolumn{5}{c}{\textbf{LADN}} \\ 
& \multirow{-2}{*}{\textbf{Deepfakes}} & \textbf{IRSE50} & \textbf{IR152} & \textbf{FaceNet} & \textbf{MobileFace} & \textbf{Average} & \textbf{IRSE50} & \textbf{IR152} & \textbf{FaceNet} & \textbf{MobileFace} & \textbf{Average} \\
\bottomrule

\multirow{3}{*}{Clean}
& UniFace   & 0.000 & 0.000 & 0.003 & 0.000 & 0.001  & 0.003 & 0.129 & 0.087 & 0.024 & 0.061 \\
& INSwapper   & 0.003 & 0.003 & 0.010 & 0.006 & 0.006  & 0.000 & 0.000 & 0.000 & 0.000 & 0.000 \\
& SimSwap  & 0.000 & 0.000 & 0.024 & 0.005 & 0.007  & 0.000 & 0.000 & 0.000 & 0.000 & 0.000 \\
\midrule



\multirow{3}{*}{TIP-IM~\cite{yang2021towards}}
& UniFace   & 0.309 & 0.681 & 0.427 & 0.347 & 0.441  & 0.725 & 0.847 & 0.680 & 0.781 & \underline{0.758} \\
& INSwapper   & 0.660 & 0.422 & 0.604 & 0.684 & \underline{0.593}  & 0.683 & 0.599 & 0.410 & 0.671 & \underline{0.591} \\
& SimSwap   & 0.775 & 0.657 & 0.754 & 0.790 & 0.744  & 0.820 & 0.799 & 0.599 & 0.820 & \underline{0.760} \\
\midrule

\multirow{3}{*}{AMT-GAN~\cite{hu2022protecting}}
& UniFace   & 0.191& 0.563& 0.793& 0.284& 0.458&0.296& 0.722& 0.716& 0.296& 0.508\\
& INSwapper & 0.218 & 0.275& 0.728& 0.359& 0.395& 0.368& 0.449& 0.695& 0.437&0.487\\
& SimSwap   & 0.486& 0.534& 0.856 & 0.580&0.614  & 0.590& 0.692& 0.766& 0.527& 0.644 \\
\midrule

\multirow{3}{*}{CLIP2Protect~\cite{shamshad2023clip2protect}}
& UniFace   & 0.505 & 0.771 & 0.757 & 0.700 & \underline{0.683} & 0.569 & 0.811 & 0.626 & 0.802 & 0.702 \\
& INSwapper  & 0.495 & 0.447 & 0.640 & 0.738 & 0.580  & 0.482 & 0.512 & 0.578 & 0.743 & 0.579 \\
& SimSwap   & 0.728 & 0.741 & 0.837 & 0.884 & \underline{0.798}  & 0.698 & 0.749 & 0.677 & 0.850 & 0.743 \\
\midrule

\multirow{3}{*}{DiffAM~\cite{sun2024diffam}}
& UniFace   & 0.095 & 0.337 & 0.582 & 0.273 & 0.322 & 0.135 & 0.353 & 0.482 & 0.296 & 0.317 \\
& INSwapper & 0.140 & 0.092 & 0.473 & 0.328 & 0.258 & 0.171 & 0.141 & 0.434 & 0.323 & 0.267 \\
& SimSwap   & 0.319 & 0.283 & 0.707 & 0.530 & 0.460 & 0.308 & 0.353 & 0.578 & 0.458 & 0.424 \\
\midrule

\multirow{3}{*}{Ours $(\lambda_{ori}{ = }1.0, \lambda_{tgt}{ = }0.5)$}
& UniFace   & 0.930 & 0.976 & 0.970 & 0.969 & \textbf{0.961} & 0.982 & 0.997 & 0.958 & 0.991 & \textbf{0.982} \\
& INSwapper   & 0.820 & 0.783 & 0.840 & 0.953 & \textbf{0.849} & 0.802 & 0.796 & 0.656 & 0.919 & \textbf{0.793} \\
& SimSwap   & 0.927 & 0.962 & 0.979 & 0.986 & \textbf{0.964} & 0.898 & 0.967 & 0.895 & 0.976 & \textbf{0.934} \\
\bottomrule

\multicolumn{12}{l}{RMT-GAN~\cite{liu2025recoverable} is excluded in this evaluation due to lack of publicly available code.}\\
\end{tabular}%
}
\end{table*}
\PP{Robustness against face-swapping deepfakes.}
We further assess robustness under face-swapping pipelines, where identity and style are explicitly disentangled. Tab.~\ref{tbl:deepfake_result} shows adversarial perturbations embedded in peripheral regions or stylistic forms (e.g, makeup), as in prior methods, often struggles to persist through the face swap, leading to reduced protection effectiveness. In contrast, $\ours$ constrains perturbations to identity-relevant facial regions and optimizes them along identity-aware axes, maintaining their effect even after identity-style disentanglement. This allows $\ours$ to consistently outperform baselines across face-swapping pipelines. Specifically, we observe substantial PSR improvements of 27.8\%, 25.6\%, and 16.6\% over the best-performing baseline on CelebA-HQ, and 22.4\%, 20.2\%, and 17.4\% on LADN against UniFace, INSwapper, and SimSwap, respectively. Notably, the largest gains against UniFace, known for its dedicated identity disentanglement module, highlight the robustness of our latent- and spatial-constrained strategies. Meanwhile, PSR scores on INSwapper are relatively lower than on other pipelines. Nonetheless, $\ours$ consistently maintains superior protection across all deepfakes and datasets, as shown in Fig.~\ref{fig:df_result}.

These results reaffirm a key limitation of impersonation-driven protection such as AMT-GAN and DiffAM: increasing similarity to a target identity does not ensure sufficient separation from the original one. Face swap makes this gap more evident, underscoring the need for dodging-oriented protection that explicitly maximizes identity separation.

\PP{Post-processing Robustness Evaluation.}
To evaluate the robustness of post-processing transformations, we apply five levels of JPEG compression, Gaussian blur, and resizing to the protected images, and evaluate cosine similarity using black-box FR models. $\ours$ consistently exhibits minimal degradation, with cosine similarity increasing by no more than 10\% under all types of distortion. Notably, in the resizing, the similarity loss remains below 1\% across all scales. This robustness stems from applying perturbations at a low-resolution stage and spatially aligning them with the original image, making them less vulnerable to downstream transformations while preserving their spectral effect. These results highlight the robustness of our method against common post-processing distortions. See Supplementary Material for visual examples and detailed results.

\subsection{Impersonation Scenario Evaluation}\label{subsec:impersonation}
We evaluate $\ours$ under the impersonation setting by fixing a target identity and disabling adaptive synthesis. As shown in Tab.~\ref{tbl:qual_impersonation} and Tab.~\ref{tbl:impersonation}, $\ours$ consistently outperforms impersonation-optimized baselines on both CelebA-HQ and LADN in terms of ASR and perceptual quality. Notably, on LADN, it achieves at least a 10.2\% absolute PSR gain over all compared methods. These results indicate that our spatially constrained masked optimization generalizes effectively to impersonation scenarios without architectural changes, surpassing baselines tailored for impersonation. Compared to the dodging setting, $\ours$ exhibits notably higher visual fidelity under impersonation, suggesting that embedding target-specific identity cues requires milder perturbations than suppressing the source identity.
\begin{table}[!htbp]
\centering
\footnotesize
\caption{Perceptual Quality under Impersonation Setting}
\label{tbl:qual_impersonation}
\resizebox{\columnwidth}{!}{%
\begin{tabular}{l||ccc}
\toprule
Method & SSIM ($\uparrow$) & PSNR ($\uparrow$) & FID ($\downarrow$)  \\
\midrule

TIP-IM~\cite{yang2021towards} &  0.884 & 30.715 & 30.858\\
AMT-GAN~\cite{hu2022protecting} & 0.787&  19.505& 34.441\\
CLIP2Protect~\cite{shamshad2023clip2protect} &  0.603&  19.354& 37.117\\
DiffAM~\cite{sun2024diffam} &  0.874& 20.526 &26.102\\
RMT-GAN~\cite{liu2025recoverable} &  0.811& 21.171 & 21.254\\
\midrule
Ours $(\lambda_{ori}{ = }0.0, \lambda_{tgt}{ = }1.0)$ &  \textbf{0.975}&  \textbf{36.809}& \textbf{13.978}\\
\bottomrule
\end{tabular}%
}
\end{table}

\begin{table*}[!htbp]
\centering
\caption{ Evaluation of ASR for impersonation (targeted) scenario at FAR@0.01. For each evaluated FR model, the other three FR models are used as surrogates to generate the protected images and the numbers in brackets indicate the improvements compared to the rates before protection.}
\label{tbl:impersonation}
\resizebox{\textwidth}{!}{%
\begin{tabular}{l||cccc|c||cccc|c}
\toprule

\multirow{2}{*}{\textbf{Method}} & \multicolumn{5}{c||}{\textbf{CelebA-HQ}} & \multicolumn{5}{c}{\textbf{LADN}} \\ 
 & \textbf{IRSE50} & \textbf{IR152} & \textbf{FaceNet} & \textbf{MobileFace} & \textbf{Average} & \textbf{IRSE50} & \textbf{IR152} & \textbf{FaceNet} & \textbf{MobileFace} & \textbf{Average} \\
\bottomrule

source  & 07.3 & 03.8 & 01.1 & 12.7 & 06.2 & 02.7 & 03.6 & 00.6 & 05.1 & 03.0 \\
MI-FGSM~\cite{dong2018boosting}  &  45.8(+38.5) & 25.0(+21.2) & 02.6(+01.5) & 45.9(+33.2) & 29.8(+23.6) & 48.9(+46.2) & 25.6(+22.0) & 06.3(+05.7) & 45.0(+39.9) & 31.5(+28.5) \\
TI-DIM~\cite{dong2019evading}  &  63.6(+56.3) & 36.2(+32.4) & 15.3(+14.2) & 57.1(+44.4) & 43.1(+36.9) & 56.4(+53.7) & 34.2(+30.6) & 22.1(+21.5) & 48.3(+43.2) & 40.3(+37.3) \\
TIP-IM~\cite{yang2021towards}  &  72.1(+64.8) & 50.0(+46.2) & 11.5(+10.4) & 79.0(+66.3) & 53.2(+47.0) & 45.3(+42.6) & 50.5(+46.9) & 15.1(+14.5) & 55.3(+50.2) & 41.6(+38.6) \\
Adv-makeup~\cite{yin2021adv}  &  17.2(+09.9) & 09.5(+05.7) & 01.4(+00.3) & 22.0(+09.3) & 12.5(+06.3) & 29.6(+26.9) & 10.0(+06.4) & 01.0(+00.4) & 22.4(+17.3) & 15.8(+12.8) \\
AMT-GAN~\cite{hu2022protecting}  &  77.0(+69.7) & 35.1(+31.3) & 16.6(+15.5) & 50.7(+38.0) & 44.9(+43.7) & 89.6(+86.9) & 49.1(+45.5) & 32.1(+31.5) & 72.4(+67.3) & 60.8(+57.8) \\
CLIP2Protect~\cite{shamshad2023clip2protect} &  81.1(+73.8) & 48.4(+44.6) & 41.7(\underline{+40.6}) & 75.2(+62.5) & 61.6(+55.4) & 91.6(+88.9) & 53.3(+49.7) & 47.9(+47.3) & 79.9(+74.8) & 68.2(+65.2) \\
DiffAM~\cite{sun2024diffam}  &  92.0(\underline{+84.7}) & 63.1(\underline{+59.3}) & 64.7(\textbf{+63.6}) & 83.5(+70.7) & 75.8(\underline{+69.6}) & 95.7(\underline{+93.0}) & 66.8(+63.2) & 65.4(+64.8) & 92.0(\underline{+86.9}) & 80.0(\underline{+77.0}) \\
\midrule
source*  &  09.0 & 04.0 & 01.6 & 19.4 & 08.5 & 10.8 & 05.7 & 06.6 & 24.6 & 11.9 \\
RMT-GAN~\cite{liu2025recoverable}  &  91.4(+82.4) & 51.8(+47.8) & 31.0(+29.4) & 93.4(\underline{+74.0}) & 66.9(+58.4) & 98.5(+87.7) & 80.8(\underline{+75.1}) & 76.9(\underline{+70.3}) & 99.4(+74.8) & 88.9(\underline{+77.0}) \\
\midrule
ours ($\lambda_{ori}{=}0.0, \lambda_{tgt}{=}1.0$)  &  96.0(\textbf{+88.7}) & 73.7(\textbf{+69.9}) & 40.7(+39.6) & 93.4(\textbf{+80.7}) & 76.0(\textbf{+69.8}) & 99.7(\textbf{+97.0}) & 91.0(\textbf{+87.4}) & 71.0(\textbf{+70.4}) & 99.1(\textbf{+94.0}) & 90.2(\textbf{+87.2}) \\

\bottomrule

\multicolumn{11}{l}{ASR values for all baseline methods are cited from CLIP2Protect, DiffAM, and RMT-GAN, which report results under a unified evaluation protocol.}\\
\multicolumn{11}{l}{* indicates that the ASR of the source image may differ from other methods due to RMT-GAN’s target selection strategy.}\\

\end{tabular}%
}
\end{table*}

\subsection{Ablation Study}~\label{subsec:ablation}
\begin{table}[!htbp]
\caption{Ablation Study for selective region (SR) and masked adversarial attack (MAA).}
\label{tbl:ablation}

\large
\resizebox{\columnwidth}{!}{%
\begin{tabular}{ccccccccc}
\toprule

\multicolumn{2}{c}{\textbf{Ablation}} & \multicolumn{4}{c}{\textbf{Protection (\%)}} & \multicolumn{3}{c}{\textbf{Perceptual Quality}}                                            
\\ \cmidrule{1-2} \cmidrule{3-6} \cmidrule{7-9}
SR & MAA & R1-O & R5-O & R1-P & R5-P & SSIM ($\uparrow$) & PSNR ($\uparrow$) & FID ($\downarrow$) \\ 
\toprule

Face Box & \checkmark  & 85.80 & 79.50 & 97.42 & 94.30 & 0.90 & 31.213 & 35.087    \\ \midrule
Non-Face& \checkmark & 0.00 & 0.00   & 8.71 & 4.84 & 0.94 & 33.498 & 29.344       \\ \midrule
Face& --  & 73.10 & 61.50 & 94.09 & 87.53 & 0.95 & 33.565 & 24.141       \\ \midrule
Face& \checkmark & 84.60 & 76.80 & 97.42 & 95.16 & 0.94 & 33.057 & 24.561 \\   
\bottomrule

\multicolumn{9}{l}{RN-O: ranked-N protection rate with the protected image and the original.}\\
\end{tabular}%
}
\end{table}






\PP{Spatial Constraints for Adversarial Perturbation.}
We investigate how perturbation placement affects protection and visual quality by disabling target-guided optimization ($\lambda_{tgt}=0$) and varying spatial constraints.

Tab.~\ref{tbl:ablation} compares four configurations: full face bounding box (baseline), non-face region, face region via post hoc masking, and our masked adversarial attack that constrains optimization to the selective face region. The baseline achieves strong protection (R1-O: 85.8\%) but degrades visual quality (FID: 35.1). Conversely, non-face perturbation yields minimal protection (R1-O: 0.0\%) despite better quality, confirming that most discriminative features lie in facial regions. Post hoc masking yields similar visual quality but leads to reduced protection (R1-O: 72.1\%).

Our approach achieves a better trade-off (R1-O: 84.6\%, FID: 24.6) by constraining gradients during optimization, not after. This enforces spatial alignment and compact perturbations, preserving both robustness and visual fidelity. These results highlight the importance of integrating spatial priors directly into the optimization process.

\PP{Target Setting Strategies for Adversarial Perturbation.}
We evaluate $\ours$ under four adversarial target strategies: Fixed, a shared target across all images adopted in most existing methods; AT-Sel, an adaptive selection approach based on landmark and identity similarity as proposed in RMT-GAN; AT-Syn, our adaptive target synthesis strategy that leverages a generative model from \citet{lee2020maskgan}; and AT-Syn(*g), a variant of AT-Syn without style conditioning, following \citet{huang2023collaborative}.

\begin{table}[!htbp]
\centering
\caption{Comparison of target strategies: geometry consistency and identity similarity with the original image.}

\label{tbl:ablation_nme}
\resizebox{\linewidth}{!}{
\begin{tabular}{c||cccc}
\toprule
& \textbf{Fixed} & \textbf{AT-Sel} & \textbf{AT-Syn(g)} & \textbf{AT-Syn} \\ 
\bottomrule
face landmarks NME$^{*}$ (mean)  & 0.074 & 0.063 &  0.033 & 0.021   \\
face landmarks NME$^{*}$ (std)   & 0.028& 0.022 &  0.016 & 0.013  \\
identity similarity (mean)  & 0.036 & 0.061 & 0.248 & 0.503  \\
\bottomrule

\multicolumn{5}{l}{NME$^{*}$: normalized mean error.}
\end{tabular}
}
\end{table}




\begin{figure}[!htbp]
    \centering
    \includegraphics[width=\linewidth]{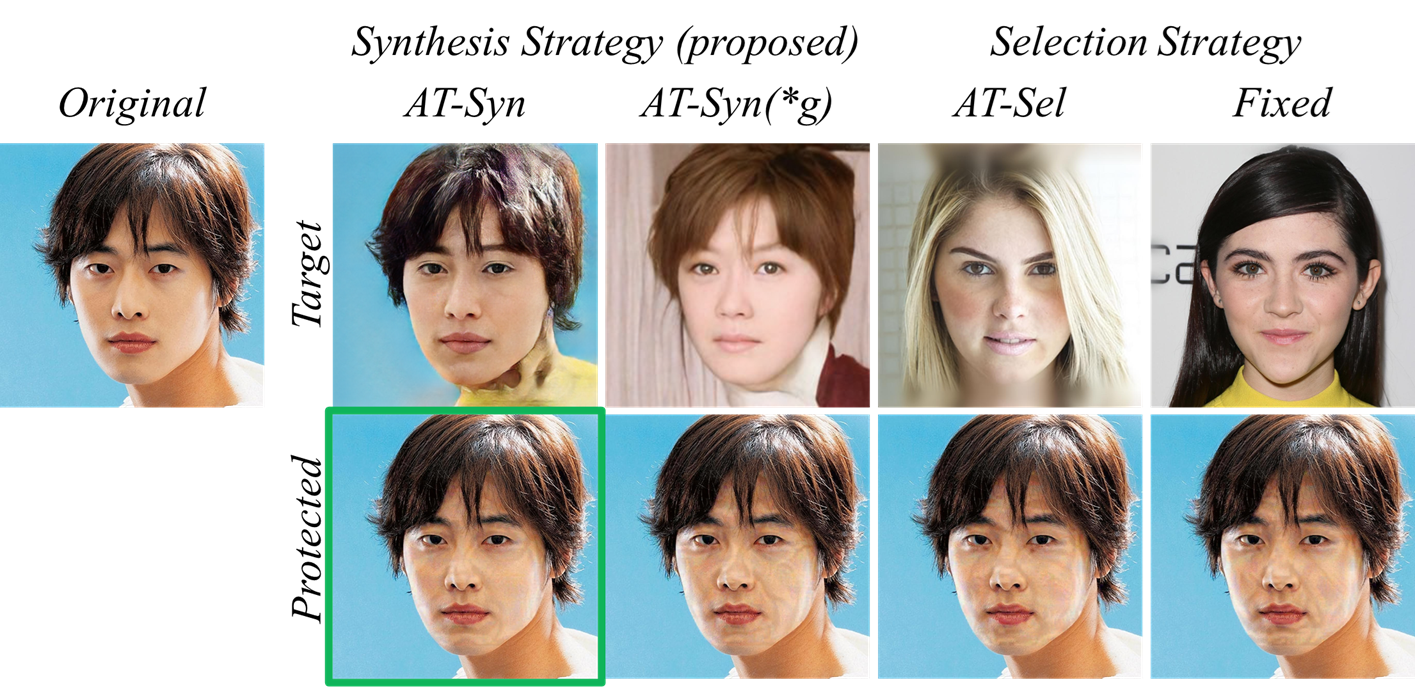}
    \caption{Sample target and protected images for dodging.}
    \label{fig:tgt}
\end{figure}
\begin{table}[!htbp]
\centering
\caption{Comparison of target strategies: protection performance (\%) and perceptual quality of protected images.}
\label{tbl:ablation_tgt}

\resizebox{\linewidth}{!}{
\begin{tabular}{c||cccccccc||cc}
\toprule
& \multicolumn{2}{c}{\textbf{IRSE50}} & \multicolumn{2}{c}{\textbf{IR152}} & \multicolumn{2}{c}{\textbf{FaceNet}} & \multicolumn{2}{c||}{\textbf{MobileFace}} & \textbf{Quality} \\ 
\multirow{-2}{*}{\textbf{Method}} & R1-P & R5-P & R1-P & R5-P & R1-P & R5-P & R1-P & R5-P & FID($\downarrow$) \\ 
\bottomrule
Fixed   &  \textbf{96.7} &\textbf{ 92.0} & 91.2 & 82.3 & 92.6 & 83.2 & 98.1 & 97.2 & 21.282 \\
AT-Sel   &  95.9 & 90.8 & 90.7 & 82.1 & 92.2 & 83.3 & 98.6 & \textbf{97.3} & 20.811  \\
AT-Syn(*g)  &  96.5 & 91.3 & 90.0 & 81.8 & 93.4 & 83.5 & \textbf{98.8} & 97.1 & 20.027 \\
AT-Syn   &  95.8 & 91.9 &\textbf{ 91.6} & \textbf{85.4} & \textbf{93.9} & \textbf{83.9} & \textbf{98.8 }& \textbf{97.3} & \textbf{18.705} \\
\bottomrule
\end{tabular}
}

\end{table}

As shown in Tab.~\ref{tbl:ablation_nme}, synthesis-based targets yield lower NME and variance, indicating superior geometry consistency, whereas selection-based targets lack such control and exhibit higher variability. In particular, AT-Syn can remain close to the source and within recognition bounds by design. Rather than generating a drastically different visual identity, AT-Syn defines a semantically meaningful direction in latent space to guide perturbation learning. Leveraging this structured signal, $\ours$ generates identity-disruptive perturbations more effectively through adversarial amplification. As shown in Tab.~\ref{tbl:ablation_tgt}, AT-Syn achieves superior perceptual quality over all alternatives, including its geometry-only variant, without sacrificing protection. Fig.~\ref{fig:tgt} further illustrates that AT-Syn yields cleaner outputs, highlighting the benefit of structure- and style-aware synthesis in guiding effective and visually coherent perturbations.

\begin{figure}[htbp!]
    \centering
    \includegraphics[width=\linewidth]{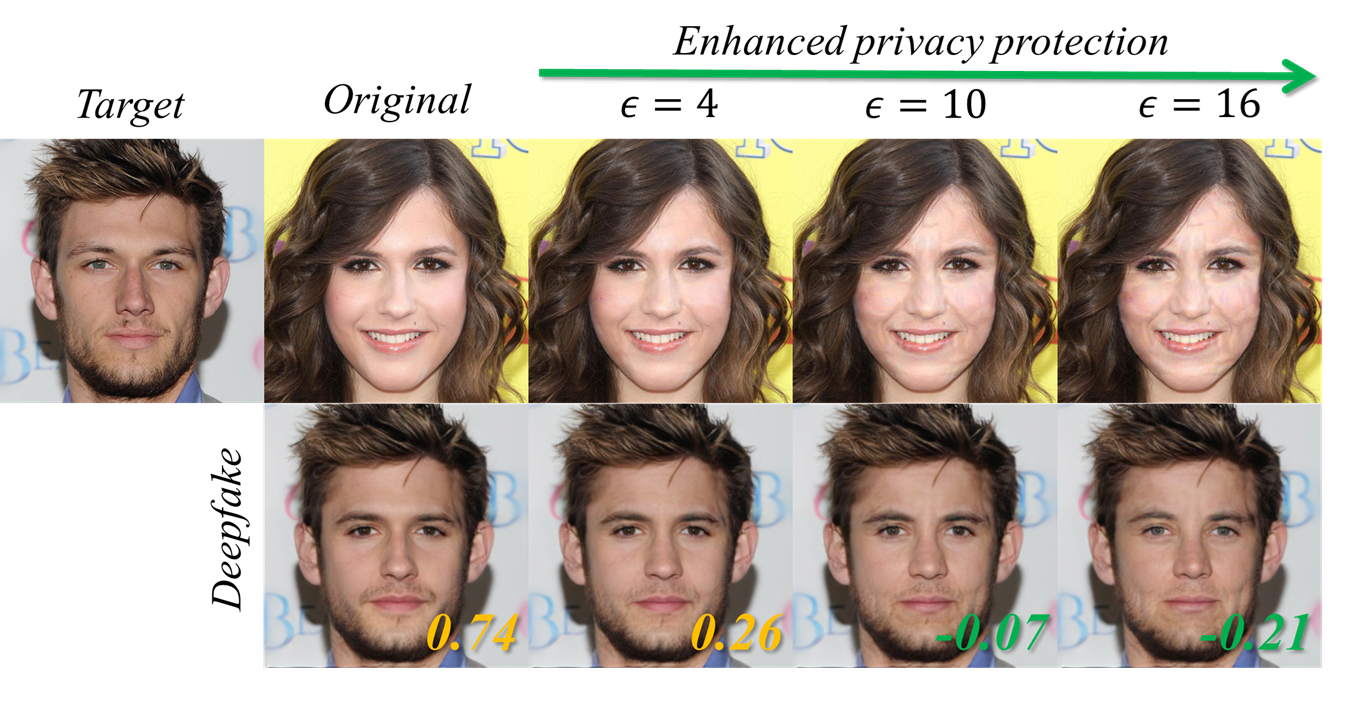}
    \caption{Deepfake results with UniFace for varying $\epsilon$ values. Numbers indicate cosine similarity on IRSE50 between deepfake and original images. Yellow/Green indicate protection failure/success at a 0.241 threshold (FAR@0.01).}
    \label{fig:eps_diff}
\end{figure}

\PP{Controllability via Perturbation Magnitude ($\epsilon$).} Fig.~\ref{fig:eps_diff} illustrates that as $\epsilon$ increases, facial distortion becomes more pronounced, yet protection against face swap improves owing to enhanced identity obfuscation. This trade-off highlights that $\ours$ allows controllable balancing between visual fidelity and protection strength. See the Supplementary Material for detailed results w.r.t. $\epsilon$ and iteration.

To further analyze robustness under imperfect masking (e.g., due to occlusion or extreme pose), 
We partition the evaluation set into two splits: $\mathcal{D}_{sts}$ (imperfect masking stress set, at least one missing face element) and $\mathcal{D}_{std}$ (reliable masking standard set, the remainder).
$\mathcal{D}_{sts}$ exhibits greater perceptual drift and weaker protection, consistent with minor spatial misalignment and slightly contaminated geometry cues that yield mildly off-identity targets and localized visible noise. Nevertheless, our method outperforms baselines because most facial regions remain covered, keeping the noise budget on salient ROIs, and the synthesized target is better aligned in style and geometry than random or fixed targets, helping preserve identity-axis guidance.
See the Supplementary for qualitative and quantitative analyses.

\section{Conclusion}
\label{sec:conclusion}
\begin{spacing}{1.0}
In this paper, we propose $\ours$, a unified face-swap deepfake protection framework for facial privacy protection that jointly constrains perturbations in latent and spatial domains. By introducing adaptively synthesized targets and enforcing region-aware updates, $\ours$ achieves robust and perceptually coherent defense against both face recognition and face-swapping deepfakes. This work highlights the importance of precise perturbation placement and guided optimization for effective identity obfuscation.
\end{spacing}

{
    \small
    \bibliographystyle{ieeenat_fullname}
    \bibliography{main}
}


\end{document}